\documentclass[10pt,twocolumn,letterpaper]{article}

\usepackage{iccv}              

\usepackage{marvosym}
\usepackage[accsupp]{axessibility}

\definecolor{iccvblue}{rgb}{0.21,0.49,0.74}
\usepackage[pagebackref,breaklinks,colorlinks,allcolors=iccvblue]{hyperref}
\usepackage{comment}
\usepackage{booktabs} 
\usepackage{colortbl}
\usepackage{graphicx}
\usepackage{xcolor}
\usepackage{multirow}
\usepackage{algorithm}
\usepackage{algpseudocode}
\newcommand{\model}{\textsc{AuroraLong}}

\def\paperID{6925} 
\def\confName{ICCV}
\def\confYear{2025}

\title{Bringing RNNs Back to Efficient Open-Ended Video Understanding}

\author{Weili Xu~$^{*}$\\
Zhejiang University\\
{\tt\small weili.23@intl.zju.edu.cn}
\and
Enxin Song~$^{*}$\\
Zhejiang University\\
{\tt\small enxin.23@intl.zju.edu.cn}
\and
Wenhao Chai~$^{\S}$\\
University of Washington\\
{\tt\small wchai@uw.edu}
\and
Xuexiang Wen\\
Zhejiang University\\
{\tt\small xuexiangwen@zju.edu.cn}
\and
Tian Ye\\
HKUST (GZ)\\
{\tt\small tye610@connect.hkust-gz.edu.cn}
\and
Gaoang Wang~$^{\dagger}$\\
Zhejiang University /    Shanghai AI Lab\\
{\tt\small gaoangwang@intl.zju.edu.cn}
}

\begin{document}

\maketitle

\renewcommand\thefootnote{}\footnote{$^{*}$ Equal Contributions. $^{\S}$ Project Lead. $^{\dagger}$ Corresponding Author.}

\begin{abstract}
The challenge of long video understanding lies in its high computational complexity and prohibitive memory cost, since the memory and computation required by transformer-based LLMs scale quadratically with input sequence length. We propose~{\normalfont\model}~to address this challenge by replacing the LLM component in MLLMs with a linear RNN language model that handles input sequence of arbitrary length with constant-size hidden states. To further increase throughput and efficiency, we combine visual token merge with linear RNN models by reordering the visual tokens by their sizes in ascending order. Despite having only 2B parameters and being trained exclusively on public data,~{\normalfont\model}~achieves performance comparable to Transformer-based models of similar size trained on private datasets across multiple video benchmarks. This demonstrates the potential of efficient, linear RNNs to democratize long video understanding by lowering its computational entry barrier. To our best knowledge, we are the first to use a linear RNN based LLM backbone in a LLaVA-like model for open-ended video understanding.
\end{abstract}
\vspace{-20pt}
\section{Introduction}
\label{sec:intro}
\begin{figure}[t]
    \centering
    \includegraphics[width=1\linewidth]{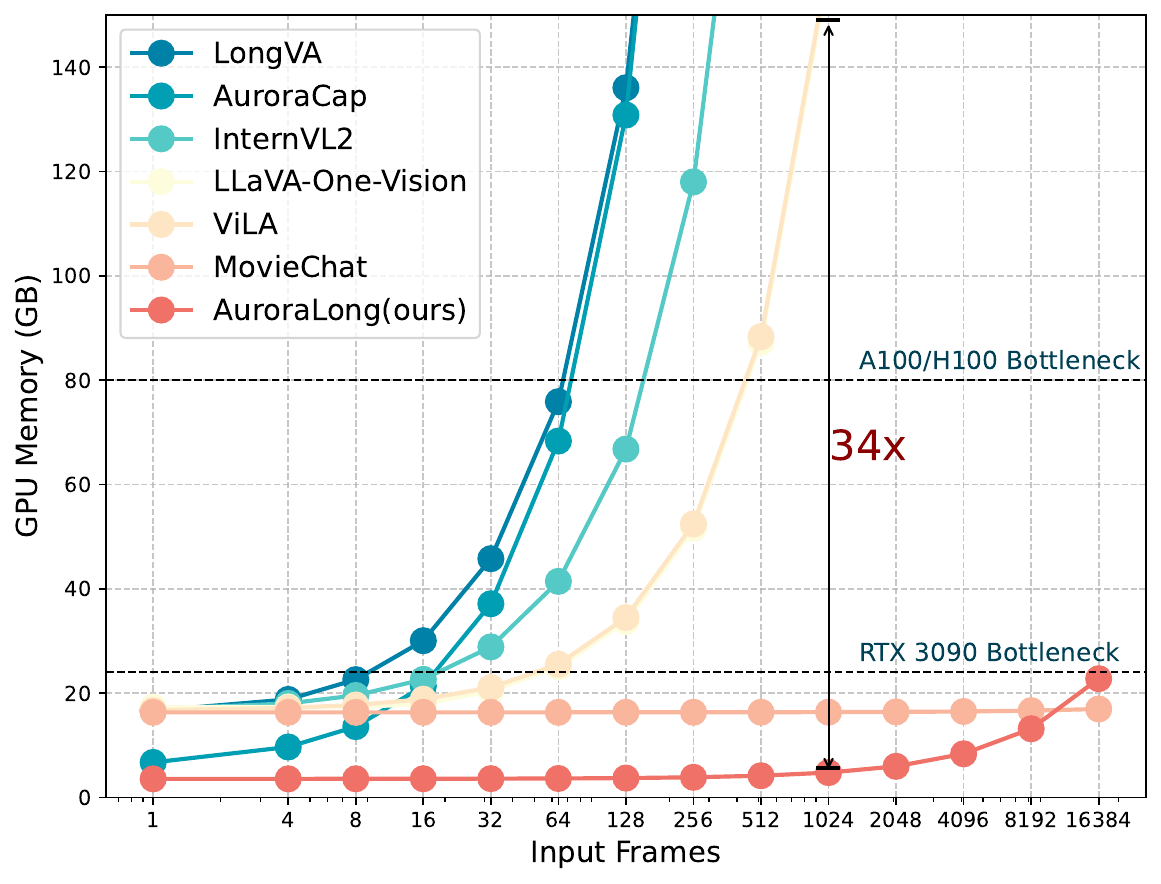}
    \caption{GPU memory cost in gigabyte (GB) (y-axis) v.s. frame number (x-axis) comparison. We test the visual-only inference of all methods. While the previous method can only support around $100$ frames of inference with A100 or H100, \model~can handle videos with over 10 thousands frames on a 24GB GPU. \model~has a $34 \times$ advantage over other methods in terms of GPU memory cost when process 1,024 frames.}
    \label{fig:intro}
    \vspace{-15pt}
\end{figure}

By integrating Transformer-based large language models (LLMs)~\cite{achiam2023gpt, touvron2023llama, yang2024qwen2} and visual extractors, large multimodal models (LMMs)~\cite{zhao2024distilling,zhou2024streaming,wang2024qwen2vl,jin2024chat, beyer2024paligemma,shang2024llava,cai2024matryoshka,cheng2024videollama} have demonstrated impressive abilities such as captioning and visual question answering. Video-based LMMs typically follow an architecture similar to LLaVA~\cite{liu2023visual}. This approach has shown promising results and scalability but faces challenges when processing longer videos with complex temporal dynamics. To comprehend rich details and dynamics in long videos, LLaVA-NeXT~\cite{liu2024llavanext} and subsequent studies demonstrate that increasing the number of sampled frames during training and inference substantially improves model performance. However, this improvement comes with considerable computational costs. As number of sampled frames increases, the computation in ViT scales linearly, as visual tokens in one frame only attend to tokens within the same frame during feature extraction, while the computation overhead in LLM scales quadratically with number of input frames due to causal self-attention mechanism, where each token attends to all previous tokens.

Currently, linear RNN large language models~\cite{peng2023rwkv, peng2024eagle, gu2023mamba, dao2024transformers} utilize linear attention~\cite{katharopoulos2020transformers, yang2025parallelizing}, which replaces the softmax attention in Transformer-based models with non-linear activation functions that are more hardware-friendly, making the memory cost to train LLMs scales linearly with regard to input sequence length. Among many linear attention variants,~\cite{peng2024eagle}~combines the parallelized training benefit of Transformers with the constant inference memory cost benefits of RNNs/LSTMs. However,~\cite{chen2024stuffed} and~\cite{zhang2024inftybench} show that linear RNN LLMs ~\cite{peng2024eagle, gu2023mamba} exhibit abnormal behavior when the context length exceeds the training length, resulting in poor long-context performance. 

Reducing the number of visual tokens has been explored in various LMMs~\cite{song2023moviechat, he2024ma, ma2024video, arif2024hired, tao2024dycoke, jie2025token}. Token Merging (ToMe)~\cite{bolya2022tome} is first introduced based on token similarity, which has proven effective in image and video classification tasks. FastV~\cite{chen2024image}, \cite{zhou2024rethinking} and \cite{zhang2024cls} prune visual tokens based on attention ranks. Similarly, to incorporate more input frames within a limited 4k context length of the pretrained Linear RNN LLM, we apply a token merging method to each layer of the vision transformer, reducing visual tokens while preserving visual information. Interestingly, our experiments show that when trained on a low token preserving ratio, linear RNN-based video LMMs tend to perform better as more visual tokens are merged, likely due to overfitting on a high visual token merging ratio.

In this paper, we present~\model, combining the simple yet efficient token merging strategy with linear RNN models by reordering the merged visual tokens, which is empirically proven to be beneficial to various video understanding tasks.
As shown in Figure~\ref{fig:intro},~\model~'s GPU memory cost remains approximately constant as the number of input frames grows. Under the 24GB memory constraint,~\model~can process up to 16K frames while being 34X more memory efficient when processing 1,024 frames. 

Our main contributions are summarized as follows:

\begin{itemize}
    \item We are the first to use a fully recurrent LLM backbone in a LLaVA-like model architecture for open-ended video QA, presenting a novel hybrid architecture that can handle long video inputs with lower memory requirement.
    \item We propose a training-free sorted visual token merge strategy to increase model throughput while retaining visual information for RNN-based large language models.
    \item Despite only trained on public data, our model performs favorably against several state-of-the-art larger LMMs across various video understanding tasks, while reducing computational complexity and memory consumption.
\end{itemize}

\section{Related Work}

\subsection{Long-form Video Understanding} 
With the develop of LLMs and LMMs~\cite{maaz2023video,zhang2023llama,li2023llamavid,zhang2023video,song2024moviechat+,fan2025videoagent, ma2023vista, xu2024pllava, li2024aria, chen2024sharegpt4video}, many recent works have broadened their application to video understanding tasks, especially for long video understanding~\cite{zhang2024internlm, zhou2024streaming, zhang2024mm, zhang2023simple, wang2024videoagent, tan2024koala, shen2024longvu, xue2024longvila, lee2024video, zeng2024timesuite, bai2025qwen25}. For long videos, the computational complexity and memory costs associated with long-term temporal connections are significantly increased, posing additional challenges. 
LongVA~\cite{zhang2024longva} processes 2000 frames (200k visual tokens) by simply extrapolating the context length of the LLM backbone and training the LMM only on short videos.
LongVILA~\cite{chen2024longvila} is trained on video inputs with 1024 frames using 256 H100 GPUs with multi-modal sequence parallelism to address the challenge of KV-cache management, which becomes a bottleneck with very long sequences.
Long video understanding is evaluated using benchmarks~\cite{yue2024movie101v2, liu2024bench, bharadwaj2024vane, du2024towards, chen2024motionllm, ataallah2024infinibench, han2023shot2story20k, xu2024motionbank, zhou2024mlvu, wang2024lvbench, fu2024video, mangalam2023egoschema, wu2024longvideobench, song2025video} typically classified as open-ended or multiple-choice questions. For open-ended questions, benchmarks like MovieChat-1K~\cite{song2023moviechat} focus on 8-minute-long movie clips.
MLVU~\cite{zhou2024mlvu} is a diverse dataset of 2593 evaluation questions on 1334 videos of varying lengths from 3 minutes to 2 hours. MVBench~\cite{li2024mvbench} is a comprehensive video QA benchmark that features 4000 multiple-choice questions that require multiple-frame input for video LMMs to answer the questions correctly.

\subsection{RNN-based Large Language Model}

Current advances in large language models (LLMs)~\cite{achiam2023gpt, touvron2023llama, yang2024qwen2} mostly focus on Transformer-based architectures, showcasing remarkable achievements across various natural language processing tasks which suffer from quadratic complexity issues in both computation and memory. Consequently, recent interest has arisen in RNN-based language models~\cite{roemmele2018automated, shen2018disan, li2018independently, feng2024were, peng2025rwkv7}. Compared to Transformer-based models, RNN-based language models inherently handle temporal sequential data, and their per-token inference cost does not increase with sequence length. However, classical RNN-based models~\cite{graves2012long, dey2017gate, schuster1997bidirectional} pose challenges in parallelization across time dimensions during training. Linear attention~\cite{katharopoulos2020transformers} replaces the softmax attention in Transformer-based models with kernel-based approximations to reduce computational cost, achieving an inference complexity of $\mathcal{O}(N)$. Some linear attention approaches~\cite{peng2024eagle, peng2023rwkv, gu2023mamba, dao2024transformers, yang2023gated, yang2024gated,zhang2024gated, de2024griffin, chai2025view} have demonstrated notable capabilities in many language processing tasks. 
Among linear attention variants, \cite{peng2024eagle}~enjoys both the benefit of transformer and RNN/LSTM, which are parallelized training and constant inference memory cost respectively. However,~\cite{chen2024stuffed} indicates that these language models may fail to extrapolate beyond pretrained context length.

\subsection{Linear Attention Models for Visual Perception} 

Previous works like Vision Mamba, and VideoMamba~\cite{zhu2024vim,duan2024vision,li2025videomamba,lu2024videomambapro, chen2024video,li2024mamba,kang2024exploring,park2024videomamba} propose light-weight image and video encoders utilizing linear attention with bi-directional scanning for traditional computer vision tasks such as image classification, action recognition, and object tracking. 
VL-Mamba~\cite{qiao2024vlmamba} mark the first attempt to use Mamba respectively as the LLM backbone for image-based LMMs for image-based visual question answering. MiniMax-VL-01~\cite{li2025minimax} is a hybrid LMM with 456B parameters that explores the scalability of lightning attention~\cite{qin2024various,qin2024lightning}, a variant of linear attention. However, no previous works use a linear RNN-based LLM as the LLM backbone for video-language modeling tasks such as open-ended video QA.

\section{Method}
\begin{figure*}[t]
    \centering
    \includegraphics[width=\textwidth]{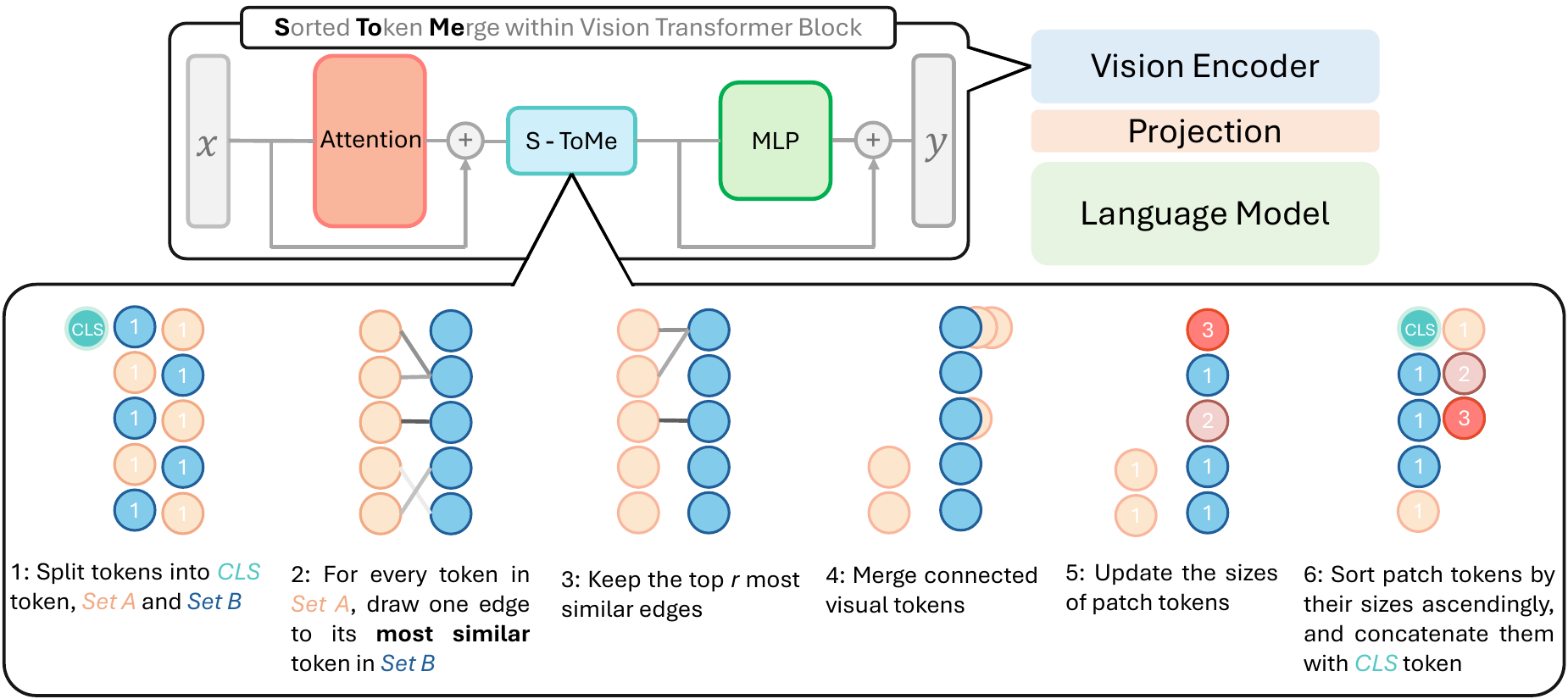}
    \caption{Visualization of the Sorted Token Merge (S-ToMe) algorithm used in~\model, whose original version is in Appendix A. }
    \label{fig:s_tome_alg}
\end{figure*}
\label{sec:method}
\subsection{Preliminaries}

\paragraph{RWKV backbone.} 

RWKV~\cite{peng2024eagle} combines the parallelizable training efficiency of Transformers with the sequential inference capabilities of RNNs. Its recurrent mechanism examines only the immediate previous token, enabling unbounded sequence lengths during inference without increasing memory requirements. RWKV-4’s core architecture computes a weighted sum of past values, modulated by a receptance vector, to efficiently facilitate information flow across time steps, which can be expressed as:
\begin{equation}
    \begin{aligned}
        \alpha_i = e^{-w} \alpha_{i-1} + e^{k_i} v_i && \beta_i = e^{-w} \beta_{i-1} + e^{k_i}
    \end{aligned}
\end{equation}
\begin{equation}
    \text{wkv}_i = \frac{e^{u + k_i} v_i + \alpha_{i-1}}{e^{u + k_i} + \beta_{i-1}}
    \label{eq:wkv_equations}
\end{equation}
where $\alpha_i$ and $\beta_i$ are recursive state variables; $k_i$ and $v_i$ are the key and value vectors at time step $i$; $w$ controls the decay rate; and $u$ is an additional learned parameter. Building upon RWKV-4, RWKV-5 uses matrix state for better expressiveness and RWKV-6 introduces data-dependent token shift with linear interpolation between input tokens. The memory requirement to train RWKV LLMs stays $O(L)$ with respect to training sample sequence length $L$.

\subsection{Method} 
\subsubsection{Network Architecture} 

We inherit the architecture of LLaVA-1.5~\cite{liu2024improved} with different choices for the vision encoder and the language model. Specifically, we use SigLIP~\cite{zhai2023sigmoid} (large-patch16-384) as the vision encoder to encode video frames and remove its final ViT layer following~\cite{wang2023cogvlm}, with a simple two-layer MLP as the cross-modal connector. We do not consider linear-attention based visual extractors since the computation overhead and memory requirement in the ViT is already $O(N)$ with respect to number of sampled frames $N$. We use RWKV-v6-Finch~\cite{peng2024eagle} as the LLM backbone for its ability to handle sequences of arbitrary length with constant memory cost. However, RNN models like RWKV lack context extension techniques like rotary position embeddings (RoPE)~\cite{su2021roformer, Qwen2VL}, necessitating the introduction of visual token merge \cite{bolya2022tome} to reduce the number of visual tokens.

\vspace{5pt}
\subsubsection{Sorted Visual Token Merge}
\definecolor{global}{RGB}{21,96,130}
\definecolor{breakpoint}{RGB}{51,0,111}

\algnewcommand{\Comment}[1]{\textcolor{blue}{\(\triangleright\) #1}}
\begin{algorithm}[t]
\caption{Sorted Visual Token Merge}\label{alg:alg}
\begin{algorithmic}
\Require Input visual tokens per frame $\mathcal{X}$ 
\Require Vision Transformer $\mathcal{V}$ with $\mathcal{N}$ layers
\Require Token Merging threshold $r$
\vspace{-10pt}
\State \For{$n$ in $\mathcal{V}[:\mathcal{N}-1)$}
\State \textcolor{global}{\# $\mathcal{X} \in [batch, tokens, channels]$}
\State $\mathcal{X} \gets \text{Attention}_{n}(\mathcal{X})$ 
\State \textcolor{global}{\# Split $CLS$ tokens and patch tokens}
\State $ CLS, \mathcal{X} \gets \mathcal{X}[:, 0 , :], \mathcal{X}[:, 1: , :]$
\State \textcolor{global}{\# Assign patch tokens to $Set~\mathcal{A}, Set~\mathcal{B}$}
\State $\mathcal{A}, \mathcal{B} \gets \mathcal{X}[: , ::2 , :], \mathcal{X}[:, 1::2 , :]$
\State $Scores \gets \text{similarity}(\mathcal{A}, \mathcal{B})$
\State \textcolor{global}{\# Get merged tokens and unmerged tokens}
\State $src, unm \gets \text{top}(\mathcal{X}, Scores, r)$
\State $dst \gets \text{merge}(src)$
\State \textcolor{global}{\# Update patch count $s$ for each token}
\State $\text{update}(dst.s)$ 
\State \textcolor{global}{\# Sort tokens by $s$}
\State $\mathcal{X} \gets \text{sort}(dst, unm)$
\State $\mathcal{X} \gets \text{concat}(CLS, \mathcal{X})$
\State $\mathcal{X} \gets \text{MLP}(CLS, \mathcal{X})$
\vspace{-10pt}
\State \EndFor
\label{alg}
\end{algorithmic}
\end{algorithm}

Despite RWKV’s~\cite{peng2024eagle} efficiency in handling inputs of arbitrary length,~\cite{chen2024stuffed} indicates that linear attention models tend to overfit to their pretrained context length. The scarcity of high-quality vision-language data compared to the vast amount of unidirectional language-only data for training RWKV makes it challenging to fine-tune the model to accommodate longer multimodal sequences. Given that RWKV~\cite{peng2024eagle} is pretrained with a context length of only 4,096 tokens, we introduce Token Merging~\cite{bolya2022tome} to merge similar visual tokens, narrowing the length gap between pretrained context and the long sequence of visual tokens. Unlike \cite{weng2025longvlm} which merges visual tokens within a video segment, we conduct token merge at a spatial level within each frame, given the consideration that when sampled at 1 FPS or lower, frame-to-frame similarity is already quite low except in static scenes, thereby obviating the need to combine visual tokens temporally.

To model visual input sequence order, Transformers utilize explicit positional embedding\cite{vaswani2017attention, ke2020rethinking, su2021roformer, touvron2023llama}, while RNNs model sequence order implicitly due to their recurrent nature. Previous works~\cite{hou2024visualrwkv, lu2024videomambapro, li2025videomamba} attempt to enhance the visual modeling capabilities of linear attention models like RWKV~\cite{peng2023rwkv, peng2024eagle} and Mamba~\cite{gu2023mamba, dao2024transformers} by bidirectionally scanning visual tokens, 
leading to additional computation overhead. Therefore, we propose a simpler, training-free visual token reordering strategy to better utilize the pretrained unidirectional textual modeling capabilities while retaining as much spatial information as possible. Specifically, as illustrated in Algorithm~\ref{alg:alg}, within each ViT layer, after merging similar visual tokens, we reorder the tokens by sorting them according to the number of visual patches they represent. We experiment with several sorting orders, and select the ascending order for its superior performance.

\vspace{-0.5em}
\subsubsection{Prompting Strategy}
\begin{figure}[t]
    \centering
    \includegraphics[width=1\linewidth]{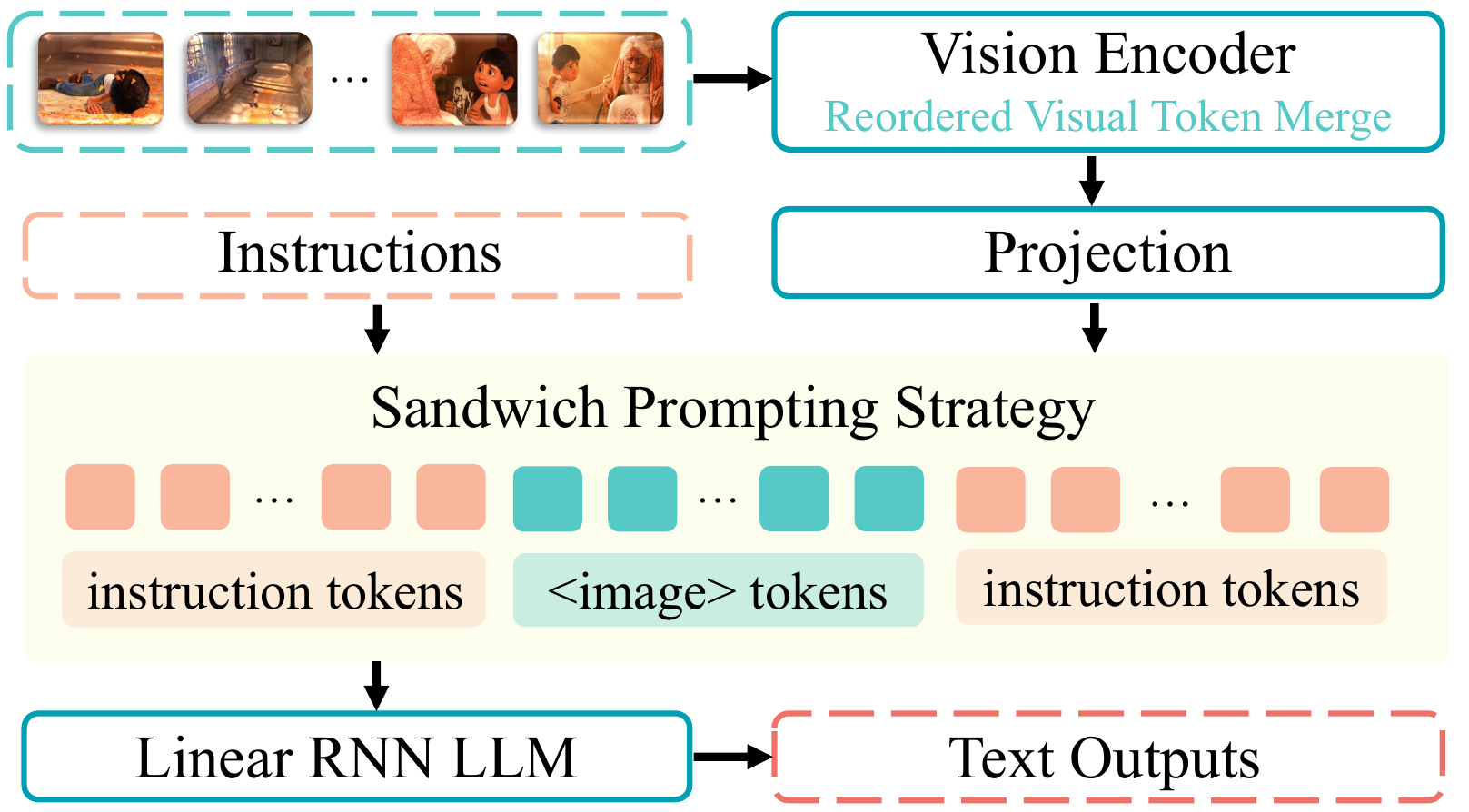}
    \caption{\model~prompting strategy overview. Following VisualRWKV~\cite{hou2024visualrwkv}, we adopt sandwich prompting strategy, which places image tokens in the middle of instruction tokens.}
    \label{fig:prompt}
\end{figure}
 Since RWKV~\cite{peng2023rwkv} and other linear RNN language models have a constant hidden state and face limitations in instruction following without careful prompts, it is crucial to employ an appropriate prompting strategy to enhance~\model's instruction following ability. Following VisualRWKV~\cite{hou2024visualrwkv}, we utilize the sandwich prompting strategy, and insert the reordered merged visual tokens between the instruction tokens as illustrated in Figure~\ref{fig:prompt}.
\subsection{Training Recipe}
Following 
\cite{chai2024auroracap}, we further adopt a three-stage training strategy, which can be noted as Pretraining stage, Vision stage and Language stage. All data we use to train~\model~are publicly available, making the experiment easy to replicate. The training data and hyperparameters used in each stage are shown in Appendix B.

\vspace{-0.5em}
\paragraph{Pretraining stage.} Similar to LLaVA~\cite{liu2024visual}, we first learn the alignment between visual features from the vision encoder and the word embedding space of RWKV~\cite{peng2024eagle}. To achieve this, we freeze the pretrained ViT and LLM, training solely the multimodal connector on image-caption pairs.

\vspace{-0.5em}
\paragraph{Vision stage.} To achieve better vision generalization, we next unfreeze the pretrained ViT while freezing the LLM during the vision stage. Note that the data we use for this stage are from various image-based computer vision tasks, which may involve labels consisting of only a few words or short phrases. Therefore, we freeze the LLM to avoid degradation in its performance as in~\cite{chai2024auroracap} and \cite{bai2025qwen25}.

\vspace{-0.5em}
\paragraph{Language stage.} Finally, we conduct end-to-end training using high-quality public data. To maintain context length similarity among samples and improve training efficiency, we distinguish the single-image data from the multiple-image samples (mainly from videos). Additionally, we set the visual token retention ratio to 0.1 so that we can feed as much input frames to~\model~as possible while further enhancing the training efficiency. We start by training with high-quality single-image data and then transit to video datasets with a lower learning rate. To improve video understanding performance, we train on video captioning samples and video question answering samples for two epochs.
\begin{figure}[t]
    \centering
    \includegraphics[width=1\linewidth]{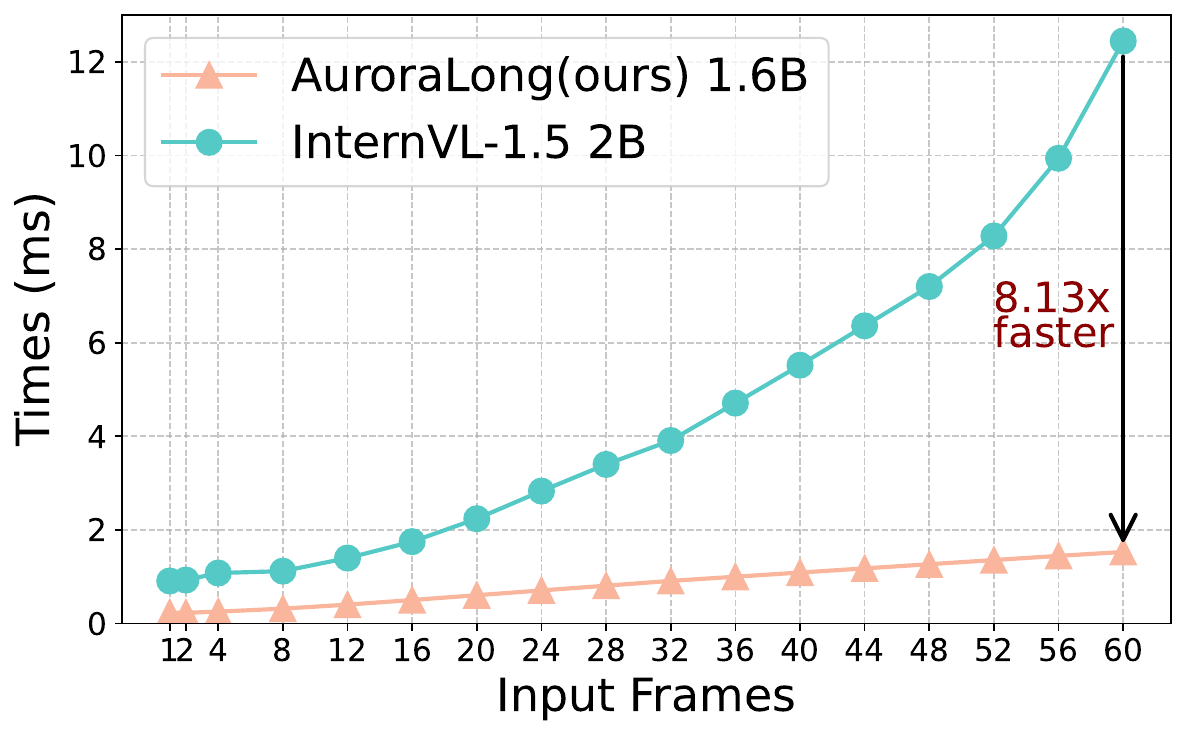}
    \caption{Compared to transformer models of similar size,~\model~requires less computation and provides lower latency.}
    \label{fig:speed}
\end{figure}

\definecolor{duration}{RGB}{253, 252, 220} 
\definecolor{ours}{RGB}{240, 113, 103}

\begin{table*}[t]
\centering
\caption{Results on short video understanding benchmarks. The best result is highlighted in bold, and the second best is underlined. We find that despite its smaller size,~\model~outperforms existing models that have much larger parameters across various short video understanding tasks. Results with * are evaluated in-house, while others are sourced from official leaderboards.
}
\resizebox{\textwidth}{!}{%
\begin{tabular}{lcc | cccccc | cc | c}
\toprule
\multirow{2}{*}{\textbf{Models}} & \multirow{2}{*}{\textbf{Size}} & \multirow{2}{*}{\textbf{\#Frame}} & \multicolumn{6}{c|}{\textbf{VDC~\cite{chai2024auroracap}}} & \multicolumn{2}{c|}{\textbf{ANet}~\cite{caba2015activitynet}} & \textbf{VATEX}~\cite{wang2019vatex} \\
& & & \textbf{Avg.} & \textbf{Short} & \textbf{Camera} & \textbf{Background} & \textbf{Main Object} & \textbf{Detailed} & \textbf{Acc.} & \textbf{Score}& \textbf{BLEU@1} \\
\midrule
Gemini-1.5-Pro & - & 1fps & \underline{41.73} & \underline{35.71} & \underline{38.68} & \textbf{43.84} & \textbf{47.32} & \underline{43.11} & - & - & -  \\
\midrule

BLIP-3-Video~\cite{xue2024xgen} & 4B & 16 & - & - & - & - & - & - & 56.9 & 3.6 & 53.2*\\
LLAMA-VID~\cite{li2025llama} & 7B & 1fps & 30.86 & 29.92  & 39.47 & 28.01 & 31.24 & 25.67 & 47.4 & 3.3 & 44.9* \\
Video-ChatGPT~\cite{maaz2023video} & 7B & 100 & 31.12 & 29.36 & 37.46 & 33.68 & 30.47 & 24.61 & 35.2 & 2.8 & 56.9* \\

LLaVA-NeXT~\cite{zhang2024llavanextvideo} & 7B & 32 & 35.46 & 30.63 & 39.73 & 36.54 & 36.54 & 33.84 & 53.5 & 3.2 & 54.8*  \\
LongVA~\cite{zhang2024longva} & 7B & 64 & 34.50 & 31.94 & 35.32 & 36.39 & 40.95 & 27.91 & - & 2.8 & 65.2* \\
ShareGPT4Video~\cite{chen2024sharegpt4video} & 8B & 16 & 36.17 & 39.08 & 33.28 & 35.77 & 37.12 & 35.62 & - & - & 56.6* \\
LLAVA-OneVision~\cite{li2024llava} & 7B & 32 & 37.45 & 32.58 & 37.82 & 37.43 & 38.21 & 41.20 & 56.6 & - & 54.2* \\
AuroraCap~\cite{chai2024auroracap} & 7B & 16 & 38.21 & 32.07 & {43.50} & {35.92} & 39.02 & {41.30} & \textbf{61.8} & \underline{3.8} & {57.1} \\
InternVL-2~\cite{chen2023internvl} & 8B & 16 & 37.72 &  33.02 &
39.08  & 37.47 & 44.16 & 34.89 & - & - & -
\\
\midrule
\model~(ours) &  2B &  1fps & \textbf{42.54} & \textbf{38.89} & \textbf{43.70} & \underline{40.26} & \underline{46.32} & \textbf{43.54} & \underline{60.0} & \textbf{4.2} & \textbf{68.5} \\
\bottomrule
\end{tabular}%
}
\label{tab:shortvideo}
\end{table*}
\section{Experiments}

In this section, we conduct both quantitative and qualitative evaluations comparing~\model~with previous methods on various video understanding tasks. We also conduct ablation studies to evaluate model performance.

\subsection{Quantitative Evaluation} 

\subsubsection{Efficiency Analysis}

As shown in Figure~\ref{fig:intro} and Figure~\ref{fig:speed}, we compare the GPU memory consumption and inference speed directly with existing leading methods. While the memory consumption of other transformer-based models increases rapidly in a quadratic manner, ~\model~consumes significantly less GPU memory, which grows linearly with respect to the number of input frames. Despite the fact that when processing videos exceeding 10,000 frames, ~\model~ requires slightly more GPU memory than MovieChat~\cite{song2023moviechat} which adopts a constant sliding window for short-term feature extraction,~\model~does not require additional memory mechanisms and consumes substantially less GPU memory when processing fewer frames. On the other hand, ~\model~ also achieves a significantly faster inference speed. In practice, when compared with InternVL-1.5 2B~\cite{chen2024far}, ~\model~ has a 34X advantage in GPU memory consumption when processing videos with 1,024 sampled frames and achieves an 8X improvement in inference speed when processing one-minute long videos at 1fps.

\subsubsection{Short Video Understanding}

We primarily conduct three tasks to assess the short video understanding capability of~\model: video question answering, video captioning, and video detailed captioning.

We conducted experiments to evaluate short video perception on multiple public datasets that provide various annotations with average video durations under 120 seconds. This includes open-ended question-answering, short captioning, and detailed captioning like VDC~\cite{chai2024auroracap}. 

For open-ended video question answering and dense captioning, we use LLM-assisted evaluation with default model choices and hyperparameter settings in LMMs-Eval \cite{lmms_eval2024, zhang2024lmmsevalrealitycheckevaluation}. Following the standard practice in VideoLLM evaluation, we report a percentage accuracy and an average score on a scale from 0 to 5. For video sparse captioning, we assess~\model~using the CIDEr (C), BLEU-4 (B@4), BLEU-1 (B@1), METEOR (M), and ROUGE-L (R) metrics on VATEX, presenting BLEU-1 scores. Additional results are provided in Appendix C.

Although the RWKV\cite{peng2024eagle} LLM backbone is pretrained only on publicly available data,~\model~exceeds Gemini-1.5-Pro on average in VDC~\cite{chai2024auroracap}, a dense captioning benchmark for short videos.

\subsubsection{Long Video Understanding}

\begin{table*}[t]
\centering
\caption{Comparison with other methods on MLVU~\cite{zhou2024mlvu} and MovieChat-1k~\cite{song2023moviechat}. Both datasets have an average video length of about 12 minutes. Results with * are evaluated in-house, while others are sourced from official leaderboards. The best result is highlighted in bold, and the second best is underlined. CTX denotes LLM pretrained context length and maximum context length for proprietary models.
} 
\vspace{-8pt}
\resizebox{\textwidth}{!}{
\begin{tabular}{l c c c @{\hspace{10pt}} | c c c c c c c c | c c }
\toprule
\textbf{Models} & \textbf{Input} & \textbf{CTX} & \textbf{Size} 
& \multicolumn{8}{c} {\textbf{MLVU}} 
& \multicolumn{2}{c} {\textbf{MovieChat-1K}} \\

& & & &\textbf{AVG} & \textbf{AR}  & \textbf{ER} & \textbf{AO} & \textbf{AC} & \textbf{TR} & \textbf{NQA} & \textbf{PQA} & \textbf{Global} & \textbf{Break} \\
\midrule
GPT4-o & 0.5fps & - & - &\textbf{54.5} & \textbf{68.8} & {47.8} &  \underline{46.2} & \underline{35.0} & \textbf{83.7} &42.9&57.1 & - & - \\
\midrule
LLAMA-VID~\cite{li2025llama} & 1 fps & 4k & 7B &18.1 & 23.1 &  11.3 & 18.6 & 15.0 &20.9&21.7&16.0 & 51.7 & 39.1\\

mPLUG-Owl-V~\cite{ye2023mplug} & 16 frm & 4k & 7B &16.7 & 15.4 & 13.2 &  14.3 & 20.0 & 25.3&6.7&22.0 & 62.9 & 44.1\\

Video-ChatGPT~\cite{maaz2023video} & 16 frm & 2k & 7B&21.2 & 17.9& 32.1& 17.1&13.3 & 17.6&28.3&22.0 & 47.6 & 48.0\\
MovieChat~\cite{song2023moviechat} & 2048 frm & 4k & 7B&16.5 & 10.3 & 15.1 &  17.1 & 15.0 & 18.7&23.3&16.0 & 62.3 &48.3\\
Video-LLAVA~\cite{maaz2023video}  & 8 frm & 4k & 7B&30.1 & 38.5 &  26.4 &  20.0 & 21.7 & 70.3	&13.3&26.0 & 55.2 & 53.1\\
LLaVA-NeXT~\cite{liu2024llavanext} & 16 frm & 8k & 7B& 27.1 & 17.9 &  26.4 &  21.4 & 16.7 & 63.7 & 13.3 & 30.0 & 45.8 & 55.2 \\
ShareGPT4Video~\cite{chen2024sharegpt4video} & 16 frm & 8k & 8B& 34.2 & 25.6 & 45.3 &  17.1 & 8.3 & 73.6 & 31.7 & 38.0 & \underline{69.0} & \underline{60.9} \\
InternVL-1.5~\cite{chen2024far} & 	16 frm & 8k & 26B& 37.9 & 51.3 &  24.5 &  14.3 & 13.3 & 80.2 & 40.0 & 42.0 & 57.7 & 61.1 \\
LongVA~\cite{zhang2024longva} & 256 frm & 224k & 7B&42.1 & 41.0 &  39.6 &  17.1 & 23.3 & \underline{81.3}	& 46.7&46.0 & 55.9 & 56.5\\
VILA-1.5~\cite{lin2023vila} & 14 frm & 276k & 40B & 46.2 & 56.4 & 35.8 &  34.3 & 11.7 & 84.7 & 38.3 & \textbf{62.0} & 57.2 & 60.1\\ 
Video-XL~\cite{shu2024video} &256 frm & 132k& 7B&46.3 & 28.2&41.5&\textbf{48.6}&31.7 & 78.0&\textbf{50.0}&46.0 & - & - \\
LLaVA-OneVision* \cite{li2024llava} & 32 frm & 132k & 0.5B & 50.3 & 58.5 & \underline{52.4} & 28.6 & 30.9 & 67.0 & 33.3 & 42.8 & - & - \\ 

Qwen2-VL* \cite{wang2024qwen2vl} & 32 frm & 132k & 2B & 48.7 & 54.7 & 47.6 & 30.9 & 28.6 & 73.8 & 40.4 & 60.5 & - & - \\
InternVL2* \cite{chen2024expanding} & 32 frm & 200k & 2B& 48.2 & 57.4 & \textbf{57.1} & 35.7 & 33.4 & 66.7 & 28.5 & 50.0 & - & - \\

\midrule
\textbf{\model~(ours)} & 48 frm & 4k & 2B & \underline{52.7} & \underline{59.5} & \textbf{57.1} & 33.2 & \textbf{42.9} & 69.0 & \underline{45.2} &\underline{61.9} & \textbf{84.0} & \textbf{64.0} \\
\bottomrule

\end{tabular}
}
\label{tab:mlvu}
\end{table*}\label{tab:longvideo}

\begin{table*}[t]
\centering
\caption{Results on MVBench \cite{li2024mvbench} whose videos primarily range
from 5s to 35s. Results with * are evaluated in-house, while others are sourced from official leaderboards. The best result is highlighted in bold, and the second best is underlined. We find that despite only being trained on public datasets,~\model~is competitive with models of similar size trained on large-scale high-quality proprietary data.
}
\vspace{-8pt}
\resizebox{\textwidth}{!}{
\begin{tabular}[t]{@{}lc|cccccccccccccccccc@{}}
\toprule
\textbf{Models} & \textbf{Size} & \multicolumn{18}{c}{\textbf{MVBench}} \\ 
 & & Avg. & UA & AC & MA & OE & ST & AL & AP & AS & CO & CI & EN & FGA & MC & MD & OI & OS & SC   \\ \midrule
Video-LLAMA~\cite{zhang2023video} & 7B & 34.1 & 39.0 & 34.0 & 32.5 & 48.0 & 43.0 & 22.5 & 25.5 & 27.5 & 40.0& 37.0 & 30.0 & 29.0 & 22.5 & 22.5 & 40.5 & 38.0 & 45.4 \\
mPLUG-Owl-V~\cite{ye2023mplug}  & 7B & 29.4 & 23.5 & 34.5 & 31.5 & 36.0 & 34.5 & 24.0 & 20.0 & 25.0 & 37.0 & 37.0 & 25.5 & 27.0 & 22.0 & 23.0 & 24.0 & 34.0 & 40.0 \\
Video-ChatGPT~\cite{maaz2023video} & 7B & 32.7 & 26.5 & 30.5 & 39.5 & 54.0 & 31.0 & 20.0 & 26.0 & 23.5 & 33.0 & 35.5 & 29.5 & 22.5 & 25.5 & 23.0 & 28.0 & \textbf{40.0} & {48.5}  \\
MovieChat~\cite{song2023moviechat}  & 7B & 33.7 & 28.0 & 42.5 & 42.5 & 39.5 & 36.0 & 26.5 & 29.0 & 33.0 & 32.5 & 32.5 & 28.5 & 31.0 & 37.5 & 27.5 & 32.0 & 35.5 & 39.5  \\
LLAVA-NeXT~\cite{liu2024llavanext} & 7B & 32.8 & 35.0 & 35.5 & 42.5 & 34.6 & 58.0 & 20.5 & 31.0 & 33.4 & 34.5 & 17.0 & 31.5 & 38.0 & 26.5 & 25.0 & 42.0 & 13.8 & 38.5 \\
LLAMA-VID~\cite{li2025llama} & 7B & 42.0 & 56.5 & 44.5 & 41.4 & 55.6 & 84.5 & 26.5 & 43.0 & 42.0 & 39.0 & 34.5 & 36.5 & 35.5 & 28.5 & 19.0 & 37.5 & 34.0 & 40.5 \\

VILA-1.5*~\cite{lin2023vila} & 40B & 42.7 & 60.0 & 41.5 & 34.5 & 50.0 & 69.5 & 36.5 & 39.5 & 40.5 & 44.0 & 40.0 & 27.0 & 33.0 & 37.0 & 27.5 & 59.5 & \underline{38.0} & 47.5 \\ 
LLaVA-OneVision~\cite{li2024llava} & 0.5B & 45.5 & 72.5 & 43.5 & 49.5 & 50.0 & 85.5 & 12.5 & 41.0 & 54.0 & 49.0 & 35.5 & 21.5 & 42.0 & 33.0 & 17.5 & 61.0 & 32.5 & 45.5\\
ShareGPT4Video*~\cite{chen2024sharegpt4video} & 8B & 47.2 & 56.5 & 34.0 & 74.5 & 81.8 & 84.5 & 34.5 & 48.0 & 45.2 & 46.0 & 51.0 & 25.0 & 35.0 & \underline{60.5} & \textbf{54.0} & 56.5 & 33.0 & 50.0 \\

LongVA* ~\cite{zhang2024longva} & 7B & {50.8} & {68.5} & \underline{{47.0}} & {56.5} & {49.5} & \underline{{89.0}} & \underline{45.0} & 58.0 & {55.6} & \underline{61.5} & {41.0} & \textbf{{39.0}} &{43.5} & {28.0} & {36.5} & \textbf{65.5} & {30.5} & {{49.0}} \\ 
InternVL-1.5* ~\cite{chen2024far} & 26B & {50.6} & \underline{{73.5}} &{27.5} & {62.5} & {44.0} & \textbf{{89.5}} & 39.3 & \underline{{61.0}} & \underline{{62.0}} & \textbf{{64.0}} & {40.5} & {34.5} & \underline{46.5} & {33.0} & {36.0} & \textbf{65.5} & {28.5} & \underline{53.0} \\

InternVL2* \cite{chen2024expanding} & 2B & 52.9 & 60.5 & 30.5 & \textbf{78.0} & \underline{79.0} & 83.5 & 31.0 & \textbf{67.0} & \textbf{72.0} & 36.0 & \textbf{55.0} & 32.0 & 38.0 & \textbf{65.5}& 32.0 & \underline{64.0} & 30.0 & 44.5\\
Qwen2-VL* \cite{wang2024qwen2vl} & 2B & \textbf{53.5} & 73.0 & 43.5 & \underline{75.5} & \textbf{82.0} & 82.0 & 12.5 & 41.0 & 54.0 & 49.0 & 35.5 & 21.5 & \textbf{48.0} & 55.0 & \underline{45.0} & 55.0  & 29.5 & 43.0\\
\midrule
\textbf{\model~(ours)} & 2B & \underline{53.2} & \textbf{75.0} & \textbf{52.0} & 65.5 & 62.5 & 87.0 & \textbf{48.0} & 47.5 & 49.5 & 47.0 & \underline{52.0} & \underline{35.0} & \underline{46.5} & 48.5 & {44.0} & 54.0 & {37.5} & \textbf{53.5} \\ 
\bottomrule
\end{tabular}
}
\label{tab:mvbench}
\end{table*}

\begin{figure*}[ht]
    \centering
    \begin{subfigure}[t]{0.24\textwidth}
        \centering
        \includegraphics[width=\textwidth]{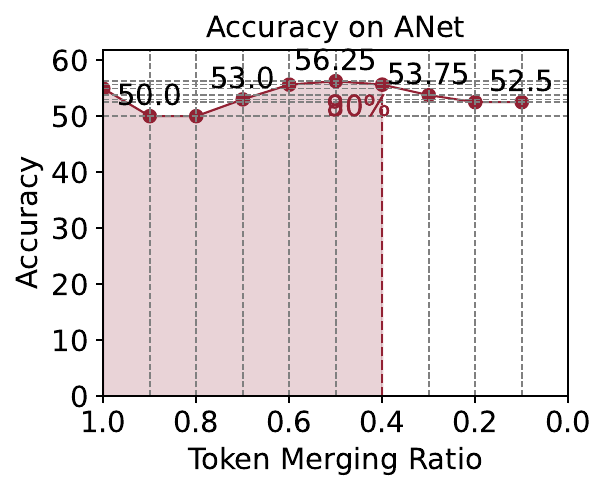}
        \caption{ANet~\cite{caba2015activitynet}}
        \label{fig:subfig1}
    \end{subfigure}
    \hfill
    \begin{subfigure}[t]{0.24\textwidth}
        \centering
        \includegraphics[width=\textwidth]{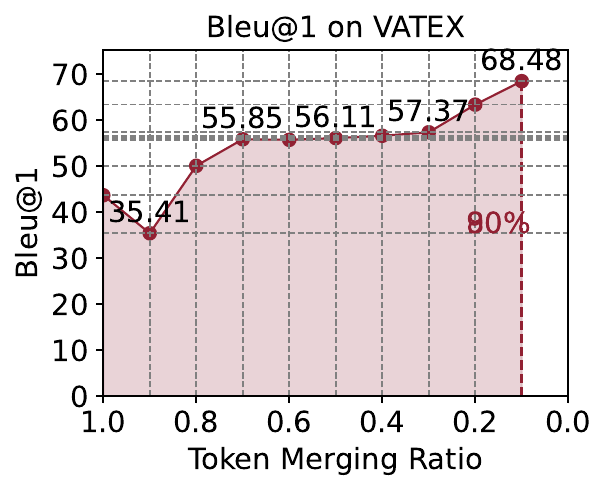}
        \caption{VATEX~\cite{wang2019vatex}}
        \label{fig:subfig2}
    \end{subfigure}
    \hfill
    \begin{subfigure}[t]{0.24\textwidth}
        \centering
        \includegraphics[width=\textwidth]{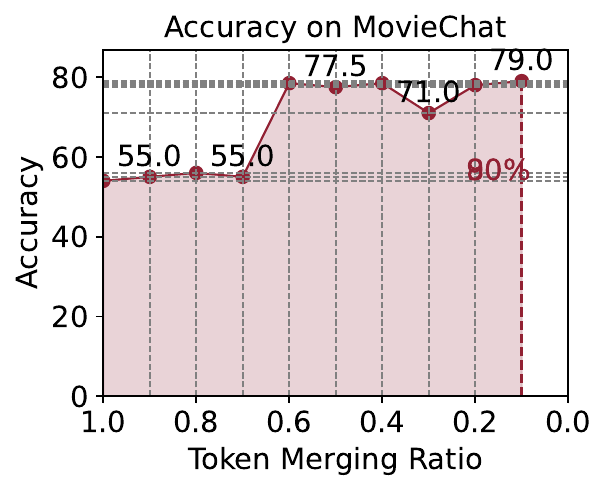}
        \caption{MovieChat-1K~\cite{song2023moviechat}}
        \label{fig:subfig4}
    \end{subfigure}
    \hfill
    \begin{subfigure}[t]{0.24\textwidth}
        \centering
        \includegraphics[width=\textwidth]{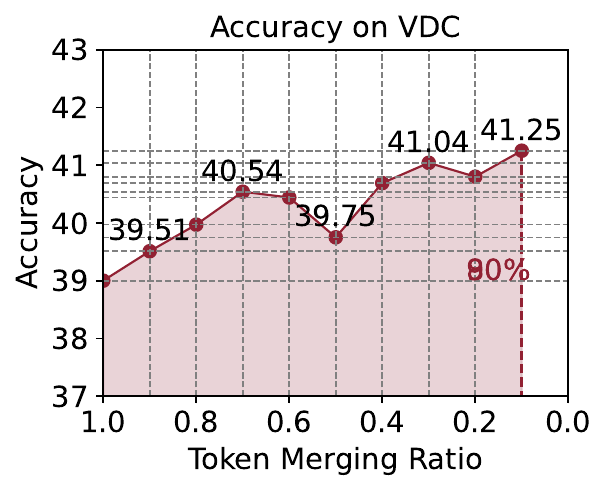}
        \caption{VDC AVG.~\cite{chai2024auroracap}}
        \label{fig:vdc-asc-subfig3}
    \end{subfigure}

    \begin{subfigure}[t]{0.24\textwidth}
        \centering
        \includegraphics[width=\textwidth]{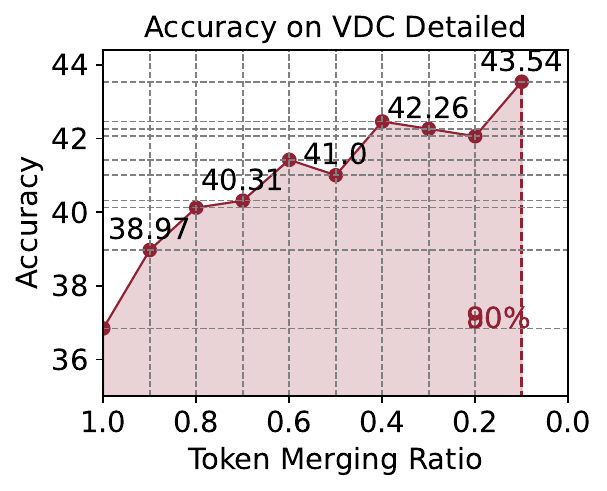}
        \caption{VDC Detailed~\cite{chai2024auroracap}}
        \label{fig:vdc1-subfig1}
    \end{subfigure}
    \hfill
    \begin{subfigure}[t]{0.24\textwidth}
        \centering
        \includegraphics[width=\textwidth]{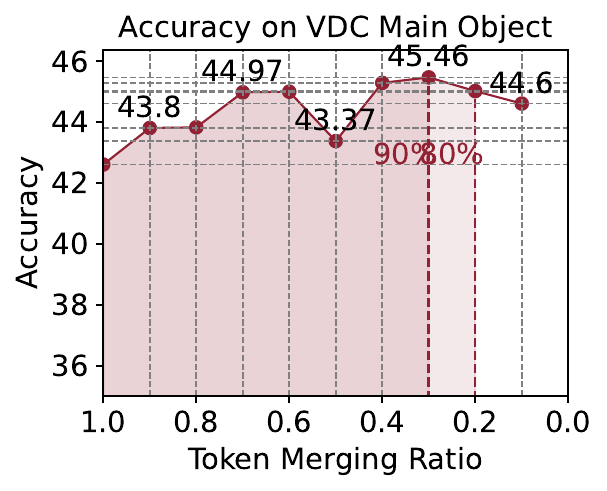}
        \caption{VDC Main Object~\cite{chai2024auroracap}}
        \label{fig:vdc2-subfig2}
    \end{subfigure}
    \hfill
    \begin{subfigure}[t]{0.24\textwidth}
        \centering
        \includegraphics[width=\textwidth]{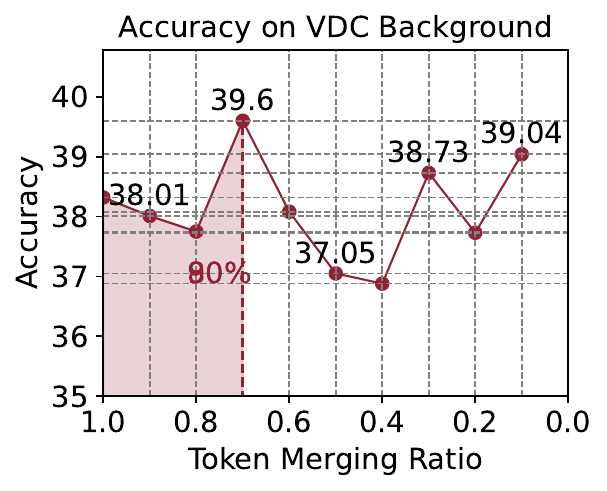}
        \caption{VDC Background~\cite{chai2024auroracap}}
        \label{fig:vdc3-subfig3}
    \end{subfigure}
    \hfill
    \begin{subfigure}[t]{0.24\textwidth}
        \centering
        \includegraphics[width=\textwidth]{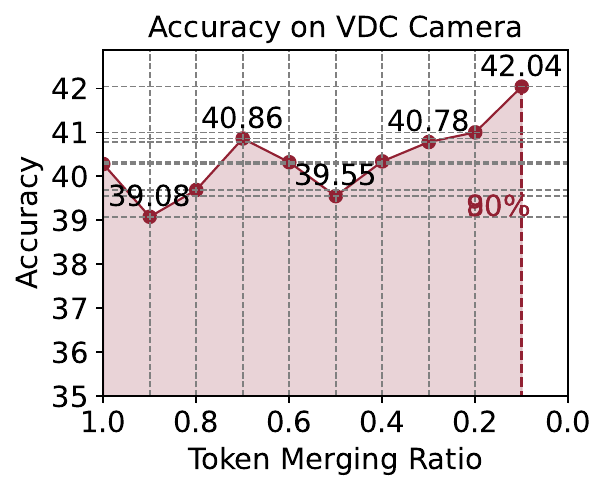}
        \caption{VDC Camera~\cite{chai2024auroracap}}
        \label{fig:vdc4-subfig4}
    \end{subfigure}
    \vspace{-8pt}
    \caption{Ablation study of token merging in short video question answering on ANet~\cite{caba2015activitynet}, short video sparse captioning on VATEX~\cite{wang2019vatex}, short video dense captioning on VDC~\cite{chai2024auroracap}, and long video question answering on MovieChat-1K~\cite{song2023moviechat}. We find that token merging significantly reduces the number of tokens while maintaining minimal performance drop, and even shows improvement in some tasks. We highlight the token merging ratio when achieving 90\% and 80\% performance with the dash line and filled area. 
    }
    \label{fig:tome_curve}
\end{figure*}

Since the RWKV LLM consumes much less memory than its Transformer counterparts when processing long input sequences, we are able to train~\model~on up to 60 input frames, which significantly enhances performance on common tasks like action counting (AC), action localization (AL) and needle QA (NQA), as is also noted in \cite{wu2024longvideobench}. Since most long video question answering tasks require understanding of multiple frames, many prior models are trained on more visual frames and use much more tokens per frame than~\model. We evaluate it on multiple long video question-answering benchmarks~\cite{song2023moviechat, li2024mvbench, zhou2024mlvu} to assess its zero-long video understanding capability. To provide a fair comparison, we follow the standard and default settings in each benchmark. To validate~\model's capability in long video understanding, we compare it with leading open weight models \cite{song2023moviechat, lin2023video, zhang2024long, li2024llava} that are up to 20 times larger than~\model~in terms of model parameter size as well as other efficient small models trained on large-scale high-quality proprietary data such as InternVL2~\cite{chen2024expanding} and Qwen2-VL~\cite{wang2024qwen2vl}. 

It is interesting that~\model~achieves comparable accuracy while consuming only about 60 tokens per frame, justifying our motivation of introducing token merge due to the spatial redundancy nature of long video understanding. Note that although~\model~was only trained on short videos within one minute and its RWKV~\cite{peng2024eagle} backbone was only trained on a context length of 4096, it still outperforms several long context Transformer-based video understanding models on long video tasks without any modifications such as adjusting $r$ in RoPE~\cite{su2021roformer} which is usually practiced on Transformer-based video understanding models. 
This generalizability aligns with the loss curve its LLM backbone shows when validated on extended textual context length that is up to 4X its pretrained context length.

\subsection{Ablation Study}
\subsubsection{Token Merging Ratio} 

As a core strategy of~\model, token merging plays a significant role in reducing the number of visual tokens, bridging the gap between the large number of video tokens and the pretrained 4k context length of RWKV LLMs. In this section, we further study how video understanding capability is influenced by token merging ratio across multiple tasks. We report the performance percentage between the highest and lowest values on the entire performance curve and identify the minimum retention thresholds for achieving 90\% and 80\% of the peak performance. As shown in Figure~\ref{fig:tome_curve}, for most tasks,~\model~reaches performance peak even with a visual token kept ratio of only 0.1. We further gather the visualization of token merging ratio on all tested video understanding tasks in Appendix A. 

Interestingly, as illustrated in Figure~\ref{fig:tome_curve}, for most video captioning tasks such as VATEX and~\cite{wang2019vatex} VDC~\cite{chai2024auroracap},~\model~'s seem to perform better at lower token retention levels. This contrasts with most Transformer-based token-reduction methods as in \cite{chai2024auroracap} and \cite{zohar2024apollo}, where performance generally declines when fewer visual tokens are retained per frame and reaches a peak performance when token kept ratio is higher than or equal to 0.5. Referring to \cite{chen2024stuffed}, we attribute this phenomenon to overfitting as the the RWKV model's recurrent state being overparameterized for the relatively short visual context length per frame in training, which is less than 60 tokens when token merge ratio is set to 0.1. Despite the overfitting tendency in spatial dimension,~\model~generalizes well in temporal dimension, handling well long videos up to 10 minutes long at zero-shot scenarios. 
More calculation details and the visualization results 
can be found in Appendix D.

\subsubsection{Input Token Order}
While most previous Linear Attention~\cite{katharopoulos2020transformers} based image-LMMs~\cite{hou2024visualrwkv} and video encoders~\cite{li2025videomamba} attempt to enhance detailed visual modeling in linear attention models by bidirectional scanning of visual tokens at the cost of increased computational complexity,~\model~simply reorders the merged visual tokens by sorting them in ascending order based on the number of tokens they combine to utilize the pretrained unidirectional textual data. 

The recurrent mechanism of RWKV~\cite{peng2024eagle} can be regarded as an implicit position encoding, which depends on sequential input token order, which is disrupted by the visual token merging process. However, the SigLIP vision encoder adds explicit positional embeddings before encoding, ensuring each token retains positional information. Thus, arranging tokens by size after merging in the vision encoder does not disrupt spatial relationships. Since we only merge visual tokens within the same frame, the temporal information is also retained. 

Additionally, we investigate how the order of merged tokens impacts performance in video understanding. In each merging operation, we merge the two most similar tokens and record the size of the merged token, i.e. total number of original tokens contained in each merged token. Before feeding the merged visual tokens into the RWKV LLM backbone, we consider three sorting strategies: no sorting (random order), sorting tokens in ascending order by size, and sorting tokens in descending order by size. 
We observe that ascending token merging performs best, as is indicated in Table~\ref{token_order}, likely because larger patches contain critical information for tasks like visual question answering, making it easier for RWKV6 to utilize its data-dependent token shifting mechanism and memorize the most critical information of each frame.

\subsubsection{Training Strategy}

In this section, we explore the alternative training strategies for the language stage of ~\model. For a fair comparison, we use the same training datasets across all settings and maintain consistent hyper-parameters. The following training settings are explored:

\begin{table}[t]
\centering
\caption{Ablation on input order for merged visual tokens within a frame, where descending order suggests tokens merged by most original tokens comes first and ascending order suggests tokens that are never merged come first among tokens of the same frame. We found that sorting merged tokens in an ascending manner brings the best performance. The best result is highlighted in bold.
}
\vspace{-8pt}
\label{token_order}
\resizebox{\linewidth}{!}{\begin{tabular}{l | c c c c}
\toprule
Token Order & ANet~\cite{yu2019activitynet} & VATEX~\cite{wang2019vatex} & VDC~\cite{chai2024auroracap} & MovieChat-1K~\cite{song2023moviechat}\\
\midrule
Random & 53.1 & 67.6& 40.9 & 76.5 \\
Descending & 55.0 & 67.0& 41.1 & 76.0 \\
Ascending & \textbf{56.3} & \textbf{68.5}&\textbf{41.3} & \textbf{78.5}\\
\bottomrule
\end{tabular}}
\end{table}

\begin{itemize}[leftmargin=1.5mm] 
\setlength{\itemsep}{2pt}
\item {\textbf{Setting $\mathbb{A}$}: To maintain a consistent number of visual tokens for the LLM, we selectively apply token merge. Specifically, for video and multi-image samples, we use a token keep ratio of 0.1. Single-image samples are left unmodified. This approach ensures a smooth transition to multi-frame training in the temporal dimension.} 
\item \textbf{Setting $\mathbb{B}$}: Inspired by Masked Autoencoders~\cite{he2022masked}, we always apply token merge (ratio 0.1) to all samples. This improves training efficiency and forces the model to learn fine-grained details from sparse tokens in single images before generalizing to multi-frame inputs.
    
\end{itemize}

\begin{figure}[t]
    \centering
    \includegraphics[width=0.99\linewidth]{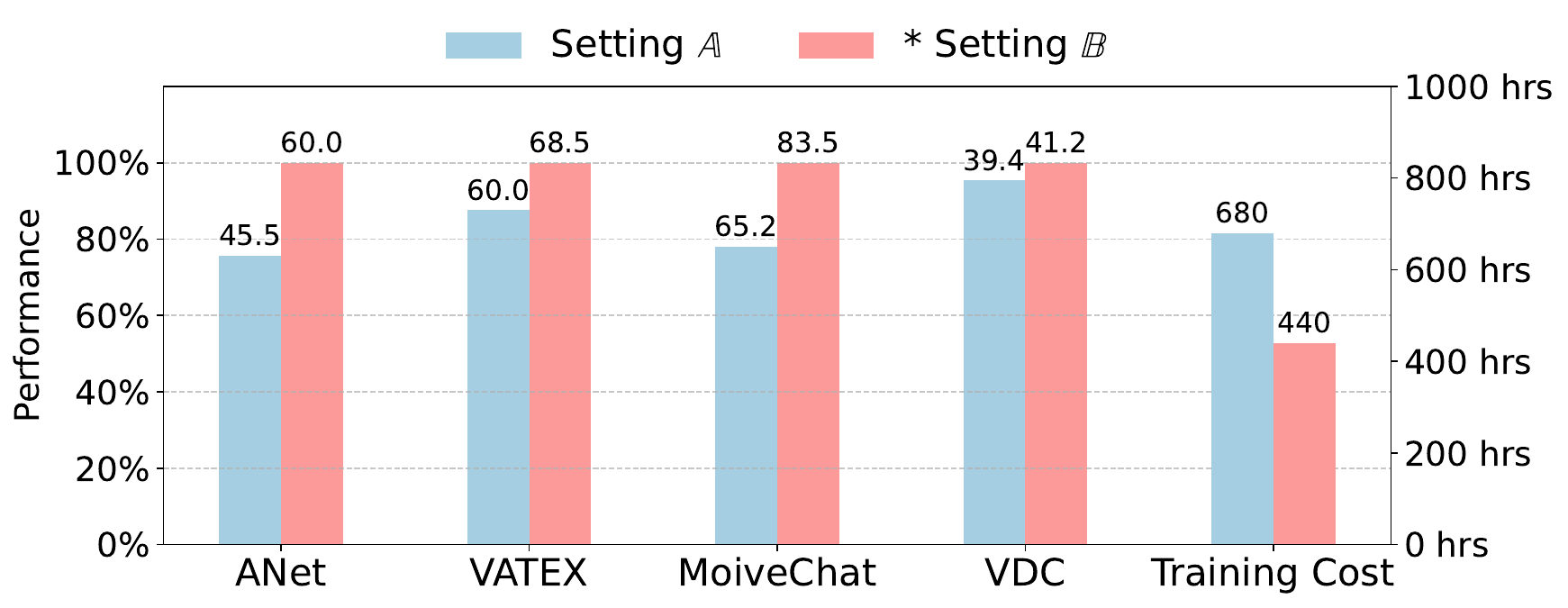}
    \vspace{-8pt}
    \caption{Comparison between different training strategy in Language stage. We take Accuracy for Question-Answering tasks and CIDEr for captioning tasks as the evaluation metric and present the performance percentage and choose Setting $\mathbb{B}$ as the final training strategy as shown with *. 
    }
    \label{fig:abl_training_strategy}
\end{figure}

We implement these two training strategies, track the training costs in A800 hours, and evaluate on various video understanding tasks. As shown in Figure~\ref{fig:abl_training_strategy}, training with setting $\mathbb{A}$ brings an extra 50\% training time overhead and leads to performance degradation across benchmarks.

\section{Limitation}
Although~\model~demonstrates impressive abilities in video understanding, it is still an early prototype with limitations, including: 1) Limited multiple-choice capability:~\model’s performance is hindered by the small size of the pretrained RWKV~\cite{peng2024eagle} model, affecting its understanding of complex multiple-choice questions.
2) Challenges in specific domains: Despite showing competitive performance on academic datasets,~\model~has limited capacity to address problems in certain areas. In the future, we will explore training with more high-quality data to further improve~\model's performance.
\section{Conclusion}

In this paper, we introduce~\model, an efficient video understanding model that leverages the linear RNN model RWKV~\cite{peng2023rwkv} as the language component. By employing a token merging strategy, we significantly reduce computational overhead without compromising performance and overcome overfitting on the training context length in linear attention variant models. We conduct extensive experiments on both short and long video understanding benchmarks, achieving improved performance with more input frames compared to advanced large multimodal models (LMMs) with larger parameters. Additionally, we carry out ablation studies to evaluate the effectiveness of the token merging ratio and the token reordering strategy we propose. The results validate the effectiveness of our proposed model and demonstrate that there is still room for improvement in applying linear RNNs to VLMs. We hope this work can serve a strong baseline in hybrid architecture for video understanding and facilitate further research in the field of non-transformer long video LMMs.

\section*{Acknowledgments}
This work is supported by the National Key R\&D Program of China (No. 2022ZD0162000), Zhejiang Provincial Natural Science Foundation of China (No. LZ24F030005, LD24F020016), and Scientific Research Foundation of Sichuan Provincial Department of Science and Technology, China (No.2024YFHZ0001).
{
    \small
    \bibliographystyle{ieeenat_fullname}
    \bibliography{main}
}

\end{document}